\documentclass[runningheads]{llncs}
\usepackage[T1]{fontenc}
\usepackage{amsmath}
\usepackage{amssymb}
\usepackage{graphicx}
\usepackage[hidelinks]{hyperref}
\usepackage{algorithm}
\usepackage{algpseudocode} 
\usepackage{xurl} 
\usepackage{url}
\usepackage{caption}
\usepackage{tabularx}
\usepackage{array}
\usepackage{makecell}
\usepackage{booktabs}
\newcolumntype{L}[1]{>{\raggedright\arraybackslash}p{#1}}
\newcommand{\code}[1]{\texttt{\path{#1}}}

\begin{document}
\title{Modernising Reinforcement Learning–Based Navigation for Embodied Semantic Scene Graph Generation}
\titlerunning{Modernising RL-Based Navigation for ESSG}
%
\author{Roman Küble\inst{1}\orcidID{0009-0002-0296-2804} \and
    Marco Hüller\inst{1}\orcidID{0009-0001-3047-8369} \and
    Mrunmai Phatak\inst{2}\orcidID{0009-0006-8035-2461}\and
    Rainer Lienhart\inst{2}\orcidID{0000-0003-4007-6889}\and
    Jörg Hähner\inst{1}\orcidID{0000-0003-0107-264X}}

\authorrunning{R. Küble et al.}
%
\institute{
    Organic Computing Group\and
    Machine Learning and Computer Vision Group\\
    University of Augsburg, Am Technologiezentrum 8,
    Augsburg, Germany}
\maketitle              
\begin{abstract}
    Semantic world models enable embodied agents to reason about objects, relations, and spatial context beyond purely geometric representations.
    In Organic Computing, such models are a key enabler for objective-driven self-adaptation under uncertainty and resource constraints.
    The core challenge is to acquire observations maximising model quality and downstream usefulness within a limited action budget.

    Semantic scene graphs (SSGs) provide a structured and compact representation for this purpose.
    However, constructing them within a finite action horizon requires exploration strategies that trade off information gain against navigation cost and decide when additional actions yield diminishing returns.

    This work presents a modular navigation component for Embodied Semantic Scene Graph Generation and modernises its decision-making by replacing the policy-optimisation method and revisiting the discrete action formulation.
    We study compact and finer-grained, larger discrete motion sets and compare a single-head policy over atomic actions with a factorised multi-head policy over action components.
    We evaluate curriculum learning and optional depth-based collision supervision, and assess SSG completeness, execution safety, and navigation behaviour.

    Results show that replacing the optimisation algorithm alone improves SSG completeness by 21\% relative to the baseline under identical reward shaping.
    Depth mainly affects execution safety (collision-free motion), while completeness remains largely unchanged.
    Combining modern optimisation with a finer-grained, factorised action representation yields the strongest overall completeness--efficiency trade-off.

\end{abstract}
\section{Introduction}
The ability of autonomous, embodied systems~\cite{embodied_ai_Duan} to create semantically rich world models in unknown, dynamic environments is a core requirement for intelligent behaviour.
From the perspective of \textit{Organic Computing} (OC)~\cite{oc1_Tomforde,oc2_Mueller,oc3_Mueller}, this represents a fundamental challenge of self-optimisation under uncertainty:
An agent must decide how to use its limited action budget to learn a policy beneficial for downstream tasks.
The efficiency of this process is largely determined by the system's ability to balance exploration and knowledge gain and to autonomously adapt its own navigation behaviour to the conditions of the environment.
The construction of such models serves as the necessary knowledge base for future OC-style \textit{self-X} properties such as \textit{self-configuration} or \textit{self-healing}.
A reliable representation of the environment is a prerequisite for the system to independently detect deviations and initiate appropriate corrective measures.

\textit{Semantic Scene Graphs} (SSGs)~\cite{Chang_2023} have established themselves as a particularly suitable representation for this purpose.
In contrast to purely geometric maps, SSGs organise objects, their attributes, and their semantic relations in a compact, graph-based structure~\cite{garg_semantics_2020}.
This enables reasoning about object identity and relations in addition to location.
However, constructing such graphs in real time requires targeted exploration and viewpoint control, a problem formalised by Li et al.~\cite{li_embodied_2022} as \textit{Embodied Semantic Scene Graph Generation} (ESSG).

In ESSG, an embodied agent explores an unknown scene from egocentric viewpoints and incrementally builds a semantic scene graph under a fixed action budget.
The pioneering approach for ESSG by Li et al.~\cite{li_embodied_2022} formulates graph construction as a \textit{Reinforcement Learning} (RL) problem, in which the agent is rewarded for discovering new objects and relations.
However, the approach has major scalability and stability limitations:
The use of the \textit{REINFORCE} algorithm~\cite{williams_simple_1992} often leads to unstable learning behaviour, while the severely restricted action discretisation limits the efficiency of exploration in complex environments.
Moreover, navigation is not treated as an explicitly optimisable component that can adapt to different utility signals.

In this paper, we address these limitations by presenting a modernised navigation component for ESSG, where navigation is understood as viewpoint selection for maximising scene-graph completeness.
Rather than proposing a new end-to-end pipeline, we isolate navigation as a self-optimising component under a fixed action budget, motivated by OC principles.
This abstraction lets us study how navigation choices affect scene-graph completeness and provides groundwork for future OC-style self-X mechanisms built on reliable semantic world models.

Concretely, we replace REINFORCE with \textit{Proximal Policy Optimization}\allowbreak\ (PPO)~\cite{schulman_proximal_2017} under identical reward shaping.
We then compare compact and larger, finer-grained discrete motion sets as well as a \textit{single-head policy over atomic actions} (SH) and a \textit{multi-head factorised policy over action components} (MH).
Factorised means the policy chooses rotation, step length, and stop decision separately rather than selecting one monolithic action ID.
In addition, we evaluate \textit{curriculum learning} (CL)~\cite{cur_lear_Narvekar}, i.e. a staged progression from simpler to more difficult control settings, and optional depth-based collision supervision.
By using an explicit \textsc{Stop} action, we enable autonomous termination that trades off information gain against budget usage.
We make the following contributions:
\begin{itemize}
    \item We show that modern policy optimisation improves SSG completeness under identical reward shaping.
    \item We characterise how discrete motion expressiveness affects completeness and behaviour under a fixed action horizon.
    \item We demonstrate that factorised high-resolution control improves optimisation behaviour and safety diagnostics over an equivalent atomic action set while maintaining comparable or higher completeness.
    \item We analyse how CL and depth-based supervision affect learning and execution across action regimes.
\end{itemize}

The remainder of this paper is structured as follows:
Section~\ref{sec:related-work} reviews relevant work on semantic exploration and SSG-based world models.
Section~\ref{sec:learning-setup} defines the state representation, action-space design, reward shaping, and policy architecture.
We then present the experimental design in Section~\ref{sec:experimental-design} and the results in Section~\ref{sec:evaluation}, discussing their implications for learning-based semantic exploration.
The paper concludes with a summary and an outlook on future research directions in Section~\ref{sec:conclusion}.

\setcounter{footnote}{0}
All components of the experimental setup, the complete training pipeline, and the paper’s appendix are publicly available in an accompanying GitHub repository\footnote{\url{https://github.com/kueblero/Modernising-RL-Based-Navigation-for-ESSG}}, enabling reproducibility and supporting future research efforts.

\section{Related Work}
\label{sec:related-work}

The construction of semantic models poses a central challenge for autonomous exploration, with SSGs proving particularly suitable for capturing objects, their attributes, and semantic relations in a unified structure~\cite{garg_semantics_2020}.
In the existing literature, these representations are often constructed incrementally during the execution of specific tasks.
For example, Tan et al.~\cite{tan_knowledge-based_2023} and Loo et al.~\cite{loo_open_2025} present approaches in which SSGs are constructed “on the fly” to solve complex goals such as searching for objects or answering visual questions in unknown environments.
In these works, however, exploration primarily serves as a secondary objective, while the optimisation of the navigation policy is focused on the respective downstream task.

In contrast, strategies for explicit semantic exploration emphasise model coverage and completeness.
Classically, heuristic or geometrically motivated methods such as \emph{frontier exploration}~\cite{auto_explo_yamauchi} or \emph{Next-Best-View} methods (NBV)~\cite{NBV_bircher} are used for this purpose.
Lang et al.~\cite{lang_selm_2025} present a current hybrid approach with SELM (\textit{Semantic Exploration and Long-term Monitoring}).
They establish the completeness of the SSG (\emph{3-DSG completeness}) as an explicit goal and use a combination of NBV planning and semantic heuristics to achieve efficient space coverage.
While SELM uses RL only for downstream monitoring tasks, the exploration itself is heuristically controlled and thus relies on predefined geometric rules.

Beyond purely geometric approaches, work in the field of \emph{Active Neural SLAM}~\cite{act_neur_slam_chaplot} shows that learning-based exploration strategies are capable of learning structural regularities of buildings and thus generating more efficient paths than classical frontier methods.
However, such approaches usually aim to optimise occupancy maps and not SSGs.

The task of the ESSG explicitly formulates the creation of a complete graph as an independent learning goal.
The approach by Li et al.~\cite{li_embodied_2022} uses RL to optimise the agent directly for the discovery of new semantic information.
In this context, navigation becomes the focus, as it directly determines which information can be integrated into the world model.
As Cartillier et al.~\cite{sem_mapnet_cartillier} point out, the efficiency of path selection is often the decisive bottleneck for the quality of the resulting semantic maps.

The choice of RL method and the design of the action space are crucial for successful navigation.
While early work often relied on simple discrete action sets, Wijmans et al.~\cite{dd_ppo_wijmans} demonstrate the superiority of PPO in terms of stability and scalability for complex navigation tasks.
We draw on these findings to address the limitations of the original REINFORCE baseline for the ESSG task.

In summary, our work differs from hybrid systems such as SELM~\cite{lang_selm_2025} by treating semantic exploration for SSG completeness as a learning-based navigation problem rather than relying on heuristic NBV/frontier control.
Relative to the ESSG baseline of Li et al.~\cite{li_embodied_2022}, we modernise the navigation policy optimisation and study how action-space resolution and factorised action parameterisations affect completeness and safety under a fixed action budget.
Following Li et al.~\cite{li_embodied_2022}, we therefore evaluate navigation primarily via SSG completeness (Node Recall) under a fixed step budget, and complement it with efficiency and safety measures such as episode length, path length, and Move Success Rate (translation actions that do not collide).
Here, Node Recall denotes the fraction of scene-graph nodes whose global soft visibility exceeds the discovery threshold by episode end.

\section{Learning Setup}
\label{sec:learning-setup}

In this section, we describe the agent-side learning setup, including the multimodal state representation, action-space design, reward shaping, and the policy architecture used for navigation.
Our setup combines multimodal perception, discrete action spaces with varying resolutions, and dense reward shaping for exploratory navigation in semantically structured environments.
The agent uses visual, graph-based, and temporal signals.
It supports either a SH action space or a factorised MH action space.

\subsection{Multimodal State Representation}
\label{subsec:state-representation}

\begin{figure}[ht]
    \centering
    \includegraphics[width=0.95\textwidth]{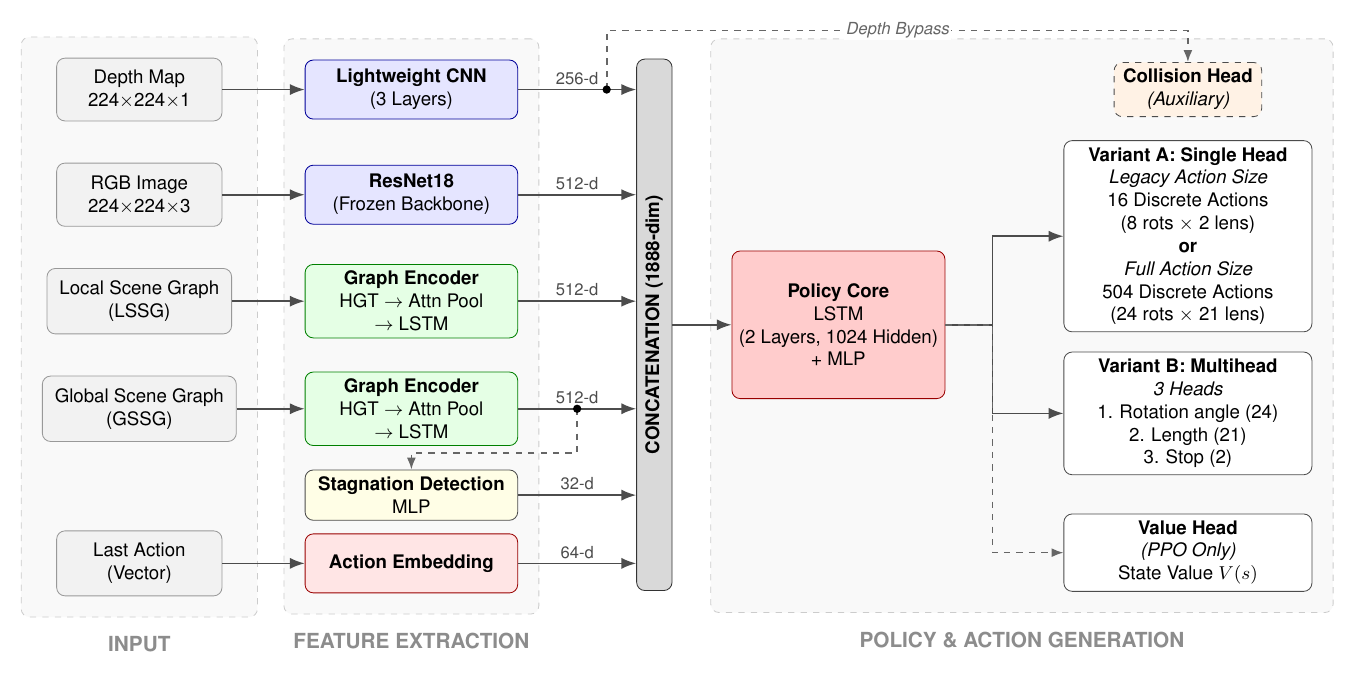}%
    \caption{Overview of the navigation model architecture used in our learning setup: multimodal encoders $\rightarrow$ LSTM policy core $\rightarrow$ action heads (single-head or multi-head) and optional collision auxiliary head.}
    \label{fig:system-overview}
\end{figure}

The agent’s state representation combines the current egocentric RGB view (and, optionally, a depth map), the previously executed action, and two SSG representations (see Fig.~\ref{fig:system-overview}) to provide comprehensive situational awareness.
This multimodal design enables the policy to jointly reason over visual observations, recent control history, and both local and accumulated semantic context.

We use two SSG representations as semantic state components:
The \textit{Local Semantic Scene Graph} (LSSG) describes the objects (nodes) and relations (edges) that can be derived from the current agent view, while the \textit{Global Semantic Scene Graph} (GSSG) incrementally aggregates this local information across the episode.
Both graphs are derived from simulation metadata (i.e., ground-truth data).
Edges (semantic relations) are generated deterministically from AI2-THOR~\cite{ai2thor} metadata using geometric and type-specific heuristics following Li et al.~\cite{li_embodied_2022}; details are provided in Appendix~\ref{app:relation-extraction}. 
The aggregation from the LSSG to the GSSG is performed using unique object IDs and thus corresponds to perfect data association; the reported results can therefore be interpreted as best-case scenarios without matching or detection errors.

Node visibility in the LSSG and GSSG is modelled via a continuous \emph{soft-visibility} heuristic that depends on object distance and apparent object size, yielding a graded visibility value in $[0,1]$ instead of a hard in-range/out-of-range decision.
We report a node as \emph{discovered} once its aggregated global soft visibility exceeds a fixed threshold.
All implementation details and parameters of the soft-visibility model and the global update rule are provided in Section~\ref{app:soft-visibility} of the appendix. 

Visual features are extracted using a ResNet18-based RGB encoder with a frozen backbone, i.e. only the newly added prediction head is trained.
When depth is enabled, we additionally use a compact depth-CNN and attach an auxiliary collision-prediction head to provide dense supervision for collision-relevant geometry.
The last action performed is converted into an action embedding.
We use a stagnation embedding derived from a change measure of the GSSG, its smoothed variant (exponential moving average), and an embedding change counter.
This embedding provides the policy with a compact progress signal to support stop decisions and reduce unproductive revisits.
Details on the calculation and the encoding can be found in Section~\ref{app:stagnation-formulas} of the appendix. 
The final state vector is obtained by concatenating all extracted features (see Fig.~\ref{fig:system-overview}).

\subsection{Action-Space Design}
\label{subsec:action-space}
We consider two policy parameterisations over the same motion primitives for navigation (see Fig.~\ref{fig:system-overview}): (i) a single-head policy over atomic actions (SH) and (ii) a factorised multi-head policy over action components (MH).
In both cases, the underlying motion resolution is defined by a discrete set of rotation angles and translation lengths.

In the SH parameterisation, each action is a single atomic choice.
We evaluate two SH configurations that differ in action-space size, angular/length resolution, and maximum translation per step.
As a compact reference, we use 16 actions corresponding to 8 rotation angles (45$^\circ$ resolution) $\times$ 2 step lengths (0.0\,m and 0.3\,m).
We additionally evaluate an enlarged SH variant with 504 atomic actions, corresponding to 24 rotation angles (15$^\circ$ resolution) $\times$ 21 lengths (0.0--2.0\,m in 0.1\,m steps).
In both SH variants, the action (rotation angle $=0$, length $=0$) is defined as \textsc{Stop} and terminates the episode.

In the MH parameterisation, the policy outputs three heads: rotation angle (24 options), length (21 options), and a binary \textsc{Stop} head, since \textsc{Stop} is atomic in SH but would otherwise require coordinated outputs across independent MH heads.
If the \textsc{Stop} head is activated, the episode terminates and the rotation/length outputs are ignored.
Otherwise, the selected rotation angle and length determine the executed motion.
In this parameterisation, (rotation angle$=0$, length$=0$) without an activated \textsc{Stop} head corresponds to an \emph{idle} action (no translation and no rotation) and does \emph{not} terminate the episode.

Actions follow a \textit{move-first} scheme where the agent first translates in the current viewing direction and then applies the selected rotation to prepare the next step.
This design ties collision risk to the currently observed forward direction, allowing the agent to use its current view to decide whether to translate at all, how far to move, or whether to rotate first to avoid an imminent collision.

For CL scenarios, we gradually expand the admissible subset of the large action space over four stages (16 $\rightarrow$ 48 $\rightarrow$ 160 $\rightarrow$ 504 actions) and mask invalid choices; details are provided in Appendix~\ref{app:curriculum}. 

\subsection{Dense Reward Shaping}
\label{subsec:reward-function}
The reward function combines multiple components to control exploration behaviour, balancing the goals of coverage, semantic completeness, and efficiency.
We define a time-dependent potential $S_t$ whose temporal difference provides a dense shaping signal, and add event-based terms for movement success ($r_t^{\text{move}}$), collisions ($r_t^{\text{coll}}$), discovery ($r_t^{\text{expl}}$), and stopping ($r_t^{\text{stop}}$).

The per-step reward is defined as
\[
    r_t
    =
    \big(S_t - S_{t-1}\big)
    +
    r_t^{\text{move}}
    +
    r_t^{\text{coll}}
    +
    r_t^{\text{expl}}
    +
    r_t^{\text{stop}} .
\]
The time-dependent shaping function $S_t$ comprises a similarity term, a viewpoint diversity term and a time penalty:
\[
    S_t
    =
    \underbrace{
        \lambda_{\text{node}}\!\left(R_{\text{node}} + \lambda_p \, P_{\text{node}}\right)
        +
        R_{\text{edge}}
        +
        \lambda_p \, P_{\text{edge}}
    }_{\text{Similarity term}}
    +
    \lambda_d \, D_t
    -
    \rho \cdot t .
\]
Here, $R_{\text{node}}$ denotes the Node Recall (proportion of nodes with visibility $\ge 0.8$), while $R_{\text{edge}}$, $P_{\text{node}}$, and  $D_t$ refer to the edge recall, the mean soft visibility across all global nodes, and the number of unique object-position pairs respectively.
We include the edge-related term $P_{\text{edge}}$ for consistency with Li et al. However, in our best-case setting edges are deterministically inferred from metadata, hence $P_{\text{edge}}=1$ and the $P_{\text{edge}}$ term does not affect the potential-based shaping reward (it cancels in $S_t - S_{t-1}$). This term becomes relevant when using a learned SSG constructor.
The diversity term counts new observations of an object from a previously unseen position, with each such observation increasing $D_t$ by 1.
The term $-\rho \cdot t$ results in a constant step penalty of $-\rho$ per time step, as rewards are given by the temporal difference $S_t - S_{t-1}$.

The shaping term derived from $S_t$ provides dense progress feedback aligned with SSG completeness and mitigates sparse-reward learning dynamics.
The event-based terms encode complementary objectives under a finite action budget: $r_t^{\text{expl}}$ rewards first-time object discoveries to encourage novelty, $r_t^{\text{coll}}$ penalises unsafe motions, $r_t^{\text{move}}$ rewards successful translation to counterbalance collision penalties and reduce rotation-only behaviour, and $r_t^{\text{stop}}$ promotes timely termination once marginal gains diminish.
The exact equations for $r_t^{\text{move}}$, $r_t^{\text{coll}}$, $r_t^{\text{expl}}$, and $r_t^{\text{stop}}$ are provided in Appendix~\ref{app:reward-terms}, Eqs.~\ref{eq:reward-coll}--\ref{eq:reward-stop}. 

We adopt the shaping parameters of $S_t$ from Li et al.~\cite{li_embodied_2022} and set the remaining coefficients via a small manual sweep on the training scenes to obtain stable learning signals and comparable reward magnitudes (i.e., preventing any single term from dominating typical episode returns).
Coefficients are not tuned on the evaluation scenes (FloorPlans~28--30) and are kept fixed across all ablation variants so that performance differences can be attributed to the ablated components at the given hyperparameters rather than reward re-weighting.
All reward hyperparameters are documented in Table~\ref{tab:hyperparameters} in the appendix. 

\subsection{Policy Architecture and Training}
\label{subsec:policy-agent}
The policy architecture uses recurrent networks to model temporal dependencies in observation sequences and follows a modular design that separates feature extraction from decision making (see Fig.~\ref{fig:system-overview}).
We support two action-output parameterisations: a SH policy that predicts a categorical distribution over atomic actions, and a MH policy with separate categorical heads for rotation angle, length, and \textsc{Stop}.
Both variants share the same feature encoders and policy core; they differ only in the factorisation of the action distribution $\pi_\theta(a_t \mid s_t)$.

The shared policy core consists of a two-layer LSTM~\cite{hochreiter_long_1997}.
Its output is fed to the respective policy head(s) and, in the PPO setting, to an additional value head for actor--critic learning.
When depth input is enabled, we further attach a collision auxiliary head that is fed directly from the depth-CNN features via a bypass, i.e., independently of the recurrent core.
This auxiliary head is trained from environment-provided motion success signals to provide dense supervision for collision-relevant geometry.
Details on the loss formulation and labelling scheme are given in Appendix~\ref{app:collision-aux}. 

We train policies either with REINFORCE~\cite{williams_simple_1992} or with PPO~\cite{schulman_proximal_2017}.
The REINFORCE objective follows the standard policy-gradient formulation with an additional entropy term and an optional collision auxiliary loss.
For PPO, we use a standard actor–critic formulation with the clipped surrogate objective and entropy regularization following Schulman et al.~\cite{schulman_proximal_2017}, and a value-function loss with value clipping as recommended by Engstrom et al.~\cite{engstrom_implementation_2020}.
Advantage estimates are computed using generalised advantage estimation~\cite{schulman_high-dimensional_2015}.
When depth input is enabled, the collision auxiliary loss is added to the PPO objective.

In the MH setup, following the action branching architecture proposed by Tavakoli et al.~\cite{tavakoli_action_2018}, the policy distribution is factorised as a product of independent head policies over rotation angle, length, and \textsc{Stop}.
Accordingly, log-prob\-a\-bil\-i\-ties and entropies decompose additively across heads.
In curriculum-learning scenarios, currently inadmissible actions (SH) or action factors (MH) are masked by setting the corresponding logits to $-\infty$.
Entropy is computed only over the admissible set, ensuring consistency between sampling and loss terms.

For the REINFORCE baseline configurations, we optionally apply imitation learning (IL) pretraining to initialize the feature encoder from expert demonstrations.
Expert data are generated offline in simulation by collecting short trajectories from multiple random start states per training scene that visit informative viewpoints and maximise SSG coverage.
At each step, we store a supervision pair consisting of the state representation and a discrete navigation action, and each trajectory ends with an explicit \textsc{Stop} action.
Details on dataset generation are provided in Appendix~\ref{app:il-dataset}. 

\section{Experimental Design}
\label{sec:experimental-design}

In this section, we describe the experimental conditions of our ablation study, including the simulation environment, the training protocol and hyperparameter selection, the considered scenarios, and the evaluation metrics.
The ablation study targets three questions:
(i) whether modern policy optimisation with PPO improves navigation efficiency and stability over REINFORCE under identical reward shaping,
(ii) how action-space resolution and factorisation affect the conversion of a limited action budget into effective exploration, and
(iii) whether CL and depth-based auxiliary supervision improve optimisation stability or final utility.

As primary performance metric, we report Node Recall, i.e., the fraction of global graph nodes whose visibility exceeds the discovery threshold ($\tau=0.8$, see Appendix~\ref{app:soft-visibility}). 
We additionally report Move Success Rate (translation actions that do not collide), episode length, and path length to characterise navigation behaviour.
Metrics are tracked during training on the training scenes and are additionally evaluated on held-out scenes by averaging over multiple episodes per scene.
Definitions and thresholds are aligned with our graph construction and reward shaping (see Appendix~\ref{app:reward-terms}). 

\subsection{Environment and Data Generation}
\label{subsec:env-data}
We use AI2-THOR~\cite{ai2thor}, an interactive indoor simulator with an embodied agent acting from egocentric viewpoints, as the simulation environment and accelerate training by caching simulator-derived state information for all discrete agent poses (position and orientation) in a database.
Transitions are then generated on-the-fly by applying the action kinematics and checking intermediate poses against the cached state set, yielding deterministic rollouts that match the simulator’s collision and movement constraints while avoiding simulator calls.

We use FloorPlans~1--27 for training and 28--30 for evaluation.
Each episode has a fixed 40-step budget, and a \textsc{Stop} action can terminate the episode early at any time.
SSG information is obtained directly from simulator metadata and is therefore free of sensor, segmentation, or detection noise.
As a result, the reported SSG metrics should be interpreted as a best-case setting under idealised perception, and would typically be lower when using realistic perception pipelines.

\subsection{Hyperparameters, Training Protocol and Scenarios}
\label{subsec:training-protocol}

In our setup, hyperparameters are tuned with Bayesian optimisation using Optuna~\cite{akiba_optuna_2019}.
For each high-level scenario (REINFORCE, SH16, SH504, and MH), we run an extensive search under a fixed compute budget to mitigate the risk that performance differences are driven by suboptimal tuning.
Ablation variants reuse the hyperparameters optimised for the depth/no-curriculum setting to keep optimisation conditions comparable and isolate architectural effects; consequently, some ablations may not be individually optimal.
The optimisation objective is a composite training metric that prioritises Node Recall while penalising collisions, truncations, and overly long episodes (Appendix~\ref{app:optuna-objective}). 
The search spaces and all hyperparameters are reported in the appendix (Tables~\ref{tab:optuna-search-space} and~\ref{tab:hyperparameters}). 

Training is performed for 1000 iterations (“blocks”) per scenario and repeated over five random seeds.
In each training block, we run $32$ environments in parallel and collect one rollout of length $60$ steps from each environment (rollouts may span multiple episodes due to resets upon termination), yielding a batch of $32$ trajectories with $60$ transitions each.
After rollout collection, the batch is used for one policy optimisation phase with Adam~\cite{kingma_adam_2017}: for PPO, this phase consists of $K=4$ optimisation epochs over the collected batch, while for REINFORCE, it consists of a single gradient update.

Evaluation on held-out scenes is performed every 50 training blocks.
At each evaluation checkpoint, we run 10 episodes per evaluation scene and report metrics averaged across all evaluation episodes.

We study a set of ablation scenarios designed to disentangle the main design choices in RL-based navigation for ESSG.
Concretely, we vary (i) the learning algorithm (REINFORCE vs.\ PPO), (ii) the action-space parameterisation and size (SH 16 vs.\ 504 actions; MH at matched 504 resolution), (iii) CL as a staged expansion mechanism for large discrete action spaces, and (iv) additional inputs and auxiliary supervision (IL pretraining and depth input with collision prediction via an auxiliary head).
Notably, the enlarged 504-action space increases both discretisation resolution and the maximum translation per step ($0.3\text{m} \rightarrow 2.0\text{m}$).
All scenarios share the same environment, reward shaping, and training protocol; thus, observed differences primarily reflect the ablated components.

The Baseline follows Li et al.~\cite{li_embodied_2022} and uses REINFORCE with IL pretraining and without depth input.
To isolate the effects of IL and depth within the REINFORCE framework, we include additional REINFORCE variants that selectively remove IL and/or add depth input.
These variants allow us to assess whether performance limitations of the baseline stem from the learning algorithm itself or from the absence of specific input modalities or pretraining.
The PPO scenarios are trained without IL and cover the compact SH action space as well as the enlarged 504-action spaces in both SH and MH formulations.
Depth input (with optional collision auxiliary supervision) is evaluated in selected PPO variants, to assess whether additional geometric cues improve execution safety or learning dynamics.
For the enlarged action spaces, we evaluate variants with and without CL to assess whether staged expansion affects optimisation behaviour and final performance.
An  overview of all scenarios is provided in Appendix~\ref{app:scenarios} (Table~\ref{tab:scenarios}). 

\section{Evaluation}
\label{sec:evaluation}

In this section, we present quantitative results on the evaluation scenes and qualitative trajectory analyses.
We then discuss the implications of these findings for utility-driven navigation in ESSG under a fixed action budget.
Scenario labels are defined in Table~\ref{tab:scenarios} in the appendix. 

\subsection{Quantitative Results}
\label{subsec:eval-quant-heldout}

We measure SSG completeness via \textit{Node Recall} on the evaluation scenes (FloorPlans~28--30) and further evaluate optimisation dynamics via \textit{Episodic Return} and execution safety on the training scenes.
Episodic Return denotes the cumulative sum of the per-step shaped reward $r_t$ (see Section~\ref{subsec:reward-function}).
Execution safety is quantified via \textit{Move Success Rate}, i.e., the fraction of executed translation actions that succeed without collision.
Pure Rotations, \textsc{Stop} actions, and idle actions in the MH setting are excluded from this metric since they do not involve translation and thus do not carry collision risk.

\begin{figure}[htbp]
    \centering
    \includegraphics[width=\textwidth]{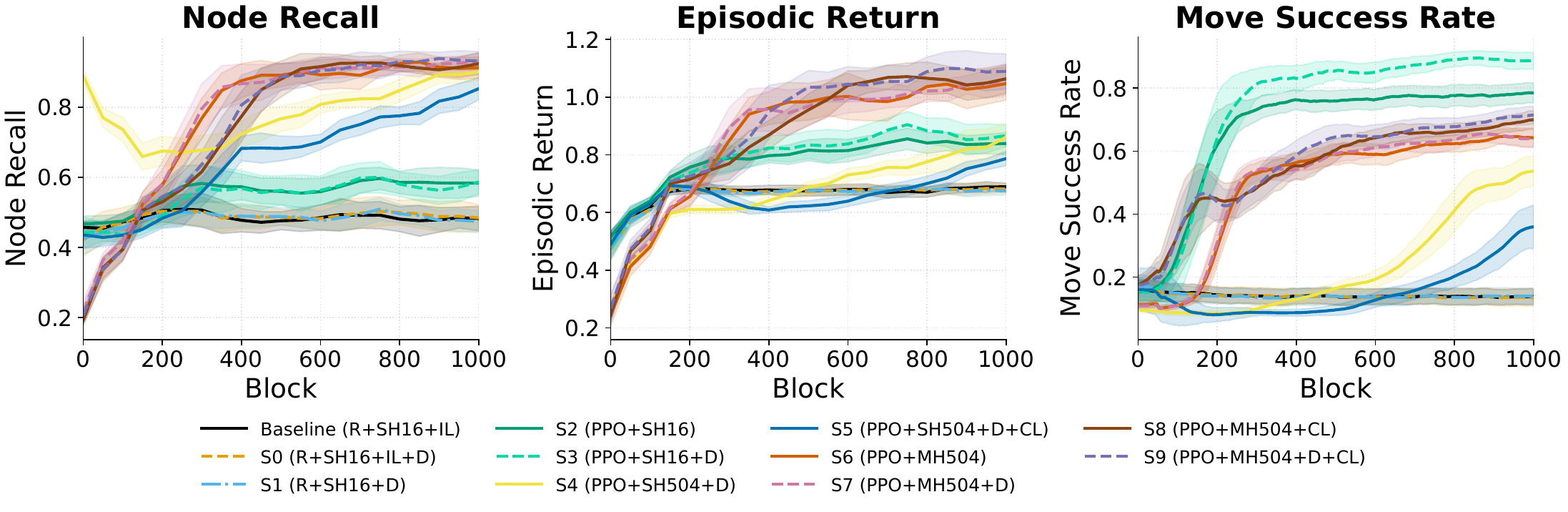}
    \caption{Learning curves over training blocks.
        Node Recall and Episodic Return are evaluated on the held-out FloorPlans~28--30 every 50 blocks, while Move Success Rate is computed on the training scenes after each block; shaded areas denote the 95\% confidence interval across seeds. SH/MH: single-head/multi-head policy; D: depth input; CL: curriculum learning; IL: imitation learning.}
    \label{fig:eval-curves}
\end{figure}

Move Success Rate was logged online on the training scenes during optimisation and is shown in Figure~\ref{fig:eval-curves} as a safety diagnostic over training.
For held-out safety, we additionally compute Move Success Rate in a post-hoc evaluation on the evaluation scenes using the final model checkpoint of each run; final held-out values are reported in Appendix~\ref{app:final-metrics}, Table~\ref{tab:final-metrics}. 
The post-hoc held-out values confirm the qualitative safety trends observed in the online diagnostic at convergence.

Across all REINFORCE variants (Baseline, S0, S1), Node Recall on the evaluation scenes remains low and largely flat, and Move Success Rate (training-scene diagnostic) stays low throughout optimisation.
Switching to PPO in the compact action space (S2) produces a rapid improvement in both Node Recall and Move Success Rate, alongside higher Episodic Return.
Adding depth input in the same compact regime (S3) further increases Move Success Rate and yields slightly higher Episodic Return, while Node Recall remains in a similar range.

Increasing action-space resolution in the atomic SH setting (S4/S5) changes the learning profile.
Both configurations improve more slowly and do not exhibit a clear Node Recall plateau within the training budget, even though they reach higher late-stage Node Recall than the compact PPO setting.
This comes with a pronounced safety cost during training:
Move Success Rate remains low for a large fraction of optimisation and temporarily drops below the REINFORCE baseline, before recovering late in training.
By the final checkpoints, Move Success Rate in SH504 is still below the other PPO scenarios, indicating that the large atomic action set is harder to learn and use safely under the fixed budget.

In the factorised MH regime (S6–S9), policies reach the highest Node Recall plateaus and exhibit high Move Success Rates in the training-scene diagnostic.
The Move Success Rate is substantially higher than the atomic high-resolution setting (S4/S5), though still typically below the compact PPO variants (S2/S3).
Depth has only marginal effects in this regime (S6 vs.\ S7), while CL primarily improves execution safety:
MH+CL (S8/S9) attains similar final Node Recall to MH without CL while achieving higher Move Success Rate earlier and at convergence, which is reflected in the highest Episodic Returns.

\begin{figure}[htbp]
    \centering
    \includegraphics[width=0.9\textwidth]{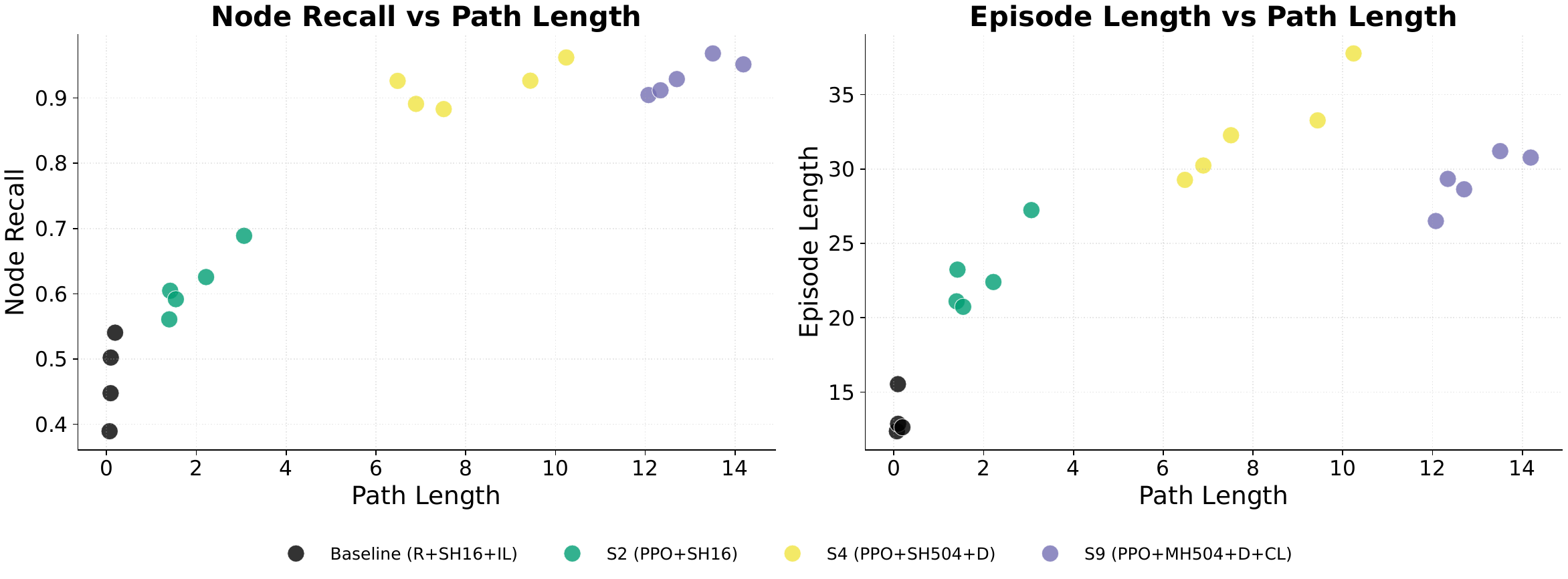}
    \caption{Evaluation scatter plot relating traversed path length to Node Recall and episode length; each point represents one trained run (one seed) at the final evaluation checkpoint.}
    \label{fig:eval-scatter}
\end{figure}

To complement the trends above, Figure~\ref{fig:eval-scatter} characterises late-stage behaviour by relating traversed path length to achieved Node Recall and episode length at the final evaluation checkpoint.
It shows that the Baseline achieves very low path lengths despite a moderate episode length, corresponding to low Node Recall.
In contrast, PPO with the compact action space (S2) reaches higher path lengths at slightly higher episode lengths and achieves higher Node Recall.
Moving from SH16 to SH504 (S4) shifts episodes towards higher Node Recall and substantially larger traversed path lengths, typically at longer episode lengths.
Among the plotted high-resolution configurations, the MH-based setup (S9) spans the highest path-length range and reaches the highest Node Recall values at shorter episode lengths and higher Move Success Rate (see Fig.~\ref{fig:eval-curves}).
Overall, higher Node Recall predominantly occurs at higher path lengths, while differences at similar path lengths indicate additional effects of step usage and action selection.

\subsection{Qualitative Trajectory Analysis}
\label{subsec:eval-qualitative}

\begin{figure}[htbp]
    \centering
    \setlength{\textwidth}{0.9\textwidth}
    \begin{minipage}{0.32\textwidth}
        \centering
        {\small\textit{FloorPlan 28}\par}
        \vspace{1pt}
        \includegraphics[width=\linewidth]{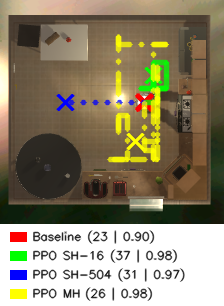}
    \end{minipage}
    \begin{minipage}{0.32\textwidth}
        \centering
        {\small\textit{FloorPlan 29}\par}
        \vspace{1pt}
        \includegraphics[width=\linewidth]{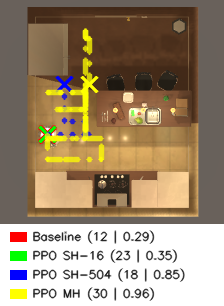}
    \end{minipage}
    \begin{minipage}{0.32\textwidth}
        \centering
        {\small\textit{FloorPlan 30}\par}
        \vspace{1pt}
        \includegraphics[width=\linewidth]{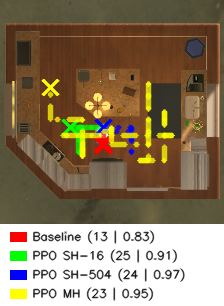}
    \end{minipage}
    \caption{Representative trajectories on evaluation scenes (FloorPlans~28--30).
        Legends report episode length and Node Recall for the shown episodes.
        All plotted scenarios except the Baseline use depth input.}
    \label{fig:qual-trajectories}
\end{figure}

To complement the quantitative results, we visualise representative trajectories to illustrate qualitative differences in navigation behaviour that are not fully captured by aggregate metrics.
Figure~\ref{fig:qual-trajectories} shows one example trajectory per evaluation scene, while additional examples across scenarios are provided in Figure~\ref{fig:qual-trajectories-app} in the appendix. 

Across the examples, the REINFORCE Baseline frequently exhibits little to no translational movement, with trajectories dominated by in-place rotations and short local motion segments.
PPO in the compact action space (SH16) shows occasional longer translations but still contains pronounced local phases and, in several cases, episodes that end by truncation at the 40-step limit rather than by \textsc{Stop}.
In the enlarged SH504 setting, trajectories are predominantly composed of longer translation segments with comparatively few direction changes, i.e., they appear more straight-line and corridor-like than in the MH setting.
By contrast, MH yields more spatially distributed trajectories with more frequent direction changes, resulting in broader coverage of the accessible area.

\subsection{Discussion}
\label{subsec:eval-discussion}

The following discussion relates the ablation results to the guiding questions defined in Section~\ref{sec:experimental-design}:
(i) the effect of the algorithm,
(ii) the role of action-space size and factorisation,
and (iii) the impact of curriculum learning and auxiliary depth-based supervision.

\paragraph{Algorithmic Modernisation.}
Replacing REINFORCE with PPO leads to a qualitative shift in optimisation behaviour.
Across all settings, PPO enables stable learning and effective use of the action budget, whereas REINFORCE remains trapped in degenerate behaviours despite identical reward shaping.
That neither imitation learning nor depth input improves REINFORCE suggests the main limitation lies in the algorithm’s optimisation dynamics.
The persistence of this gap on evaluation scenes further suggests PPO improves generalisation rather than exploiting training-specific regularities.

\paragraph{Action-Space Size and Motion Expressiveness.}
Increasing action-space resolution and step-range expands the set of reachable viewpoints within the same action budget, which directly supports higher SSG completeness in a finite-horizon setting.
However, this expressiveness comes with a trade-off: the larger decision space makes optimisation harder and slows convergence, since useful behaviours occupy a smaller fraction of the policy’s output space and require finer credit assignment.
This is also reflected in the scatter visualisation (see Fig.~\ref{fig:eval-scatter}), where enlarged action spaces reach higher Node Recall at comparable episode lengths by converting steps into longer effective traversal.
The absence of a clear Node Recall plateau in the enlarged SH setting suggests that the larger atomic action space is substantially harder to learn within a fixed budget, highlighting the need for more structured action representations beyond simply increasing resolution.

\paragraph{Factorised vs.\ Atomic Action Spaces.}
In large discrete action spaces, factorising the policy into separate heads for motion components improves controllability compared to selecting a single atomic action, and reaches strong performance markedly earlier in training.
In SH504, learning can collapse onto a small subset of high-utility atomic actions (often including 90$^\circ$ turns), since each rotation–translation combination must be learned separately.
MH mitigates this by decomposing the decision, enabling more effective reuse and recombination of learned motion components.
Consistent with this, the scatter and trajectory visualisations jointly indicate that factorisation yields more spatially distributed coverage (more direction changes), whereas SH504 tends towards straighter, corridor-like traversal, suggesting less effective utilisation of the high-resolution atomic action set.

\paragraph{Curriculum Learning (CL) Effects.}
CL depends on the action-space parameterisation.
In the atomic SH504 setup, staged expansion degrades optimisation within the fixed budget, yielding slower progress and lower returns than training directly in the full action space.
In contrast, in the factorised MH setup, CL primarily changes execution behaviour rather than achievable completeness.
The main benefit is improved safety, with higher Move Success Rate early in training and at convergence, while final Node Recall remains similar.
Overall, CL acts mainly as a safety mechanism here, introducing a trade-off between learning speed and execution reliability depending on deployment priorities.
A plausible explanation is that the early curriculum stages restrict long-range motions and thus reduce the probability of failed moves (collisions).
Consequently, Move Success Rate increases even if achievable completeness remains largely unchanged.
This interpretation is consistent with the compact action-space regime, where short step lengths exhibit high Move Success Rate (see Fig.~\ref{fig:eval-curves}).

\paragraph{Depth Input and Auxiliary Supervision.}
Depth input and collision auxiliary supervision show a setting-dependent effect.
In the compact PPO setting, adding depth substantially improves Move Success Rate and slightly increases Episodic Return, while Node Recall remains in a similar range.
In the high-resolution MH regimes, depth yields only marginal differences on the metrics, suggesting limited additional benefit beyond the remaining state inputs in this setting.

\paragraph{Limitations and Practical Implications.}

Our findings should be interpreted under several constraints.
First, SSGs are derived from simulator metadata and thus reflect an idealised perception setting without detection or data-association noise.
Second, the training and evaluation environments are small and structured, which may understate the expressiveness demands of navigation in larger or more complex settings.
Third, some configurations did not fully converge within the fixed training budget, so asymptotic differences may be underestimated.
Finally, our high-resolution action spaces jointly increase discretisation granularity and the maximum translation per step; consequently, the observed gains cannot be attributed to discretisation alone, but also reflect increased per-step reachability under a fixed step budget.

From a practical perspective, two implications stand out.
(i) Training speed and execution safety (Move Success Rate) matter for deployment on embodied platforms.
In the MH setting, policies with and without CL reach comparable completeness, but CL slows early learning.
In return, CL improves execution safety earlier and at convergence, yielding a speed–safety trade-off depending on deployment priorities.
(ii) Across our ablations, PPO combined with a factorised MH action space provides the most robust trade-off between optimisation stability, effective traversal, and completeness, making it the most promising configuration for future end-to-end ESSG systems under more realistic sensing.


\section{Conclusion}
\label{sec:conclusion}

In this work, we revisited RL-based navigation for ESSG with the goal of improving SSG completeness, safety, and navigation behaviour under a fixed action budget.
Rather than proposing a new end-to-end pipeline, we deliberately isolated navigation as an optimisable component and systematically studied how modern policy optimisation and action-space design influence SSG completeness under identical reward shaping.

Across all ablations, replacing the REINFORCE baseline with PPO led to a structural improvement in optimisation behaviour, yielding higher Episodic Return and consistently higher SSG completeness.
Expanding the discrete action space further increased coverage within the same episode horizon, while factorising the action space into interpretable motion components improved controllability and robustness in large discrete settings.
Curriculum learning was not universally beneficial, but in the multi-head setting it introduced a clear speed–safety trade-off: it improved execution safety (higher Move Success Rate), particularly early in training, while reaching comparable completeness with slower convergence.
Depth-based supervision showed a regime-dependent effect: in the compact PPO setting it substantially improved execution safety and slightly increased Episodic Return, while in the high-resolution multi-head regimes it had only marginal impact on the metrics.

From an OC perspective, our results show that navigation can be treated as a self-optimising component whose policy adapts under a fixed action budget to maximise a utility signal tied to world-model completeness.
By improving this component through modern policy optimisation and suitable action parameterisation, we strengthen a key prerequisite for OC-style self-X capabilities that depend on reliable semantic world models.

Future work will extend this foundation along two directions.
First, we will integrate the navigation module into a full ESSG pipeline with realistic perception and data association, and then evaluate the resulting best-performing configuration in larger and more complex environments to assess end-to-end scalability under noise.
Second, we will embed the navigation component into higher-level OC control loops, enabling explicit self-configuration and self-healing mechanisms that adapt to changing sensing, actuation, or utility conditions.
%
%
\bibliographystyle{splncs04}
\bibliography{sources}

@article{lang_selm_2025,
  title      = {{SELM}: {From} {Efficient} {Autonomous} {Exploration} to {Long}-{Term} {Monitoring} in {Semantic} {Level}},
  volume     = {17},
  issn       = {2379-8939},
  shorttitle = {{SELM}},
  doi        = {10.1109/TCDS.2025.3531367},
  number     = {4},
  urldate    = {2026-01-02},
  journal    = {IEEE Transactions on Cognitive and Developmental Systems},
  author     = {Lang, Fang and Qin, Yongsen and Wang, Yinchuan and Liu, Jin and Wang, Chaoqun and Song, Wei and Zhu, Qiuguo and Song, Rui},
  month      = aug,
  year       = {2025},
  pages      = {938--952}
}

@inproceedings{li_embodied_2022,
  title     = {Embodied {Semantic} {Scene} {Graph} {Generation}},
  url       = {https://proceedings.mlr.press/v164/li22e.html},
  language  = {en},
  urldate   = {2025-04-22},
  booktitle = {Proceedings of the 5th {Conference} on {Robot} {Learning}},
  publisher = {PMLR},
  author    = {Li, Xinghang and Guo, Di and Liu, Huaping and Sun, Fuchun},
  month     = jan,
  year      = {2022},
  issn      = {2640-3498},
  pages     = {1585--1594}
}

@article{tan_knowledge-based_2023,
  title     = {Knowledge-{Based} {Embodied} {Question} {Answering}},
  volume    = {45},
  copyright = {https://ieeexplore.ieee.org/Xplorehelp/downloads/license-information/IEEE.html},
  issn      = {0162-8828, 2160-9292, 1939-3539},
  doi       = {10.1109/TPAMI.2023.3277206},
  language  = {en},
  number    = {10},
  urldate   = {2026-01-10},
  journal   = {IEEE Transactions on Pattern Analysis and Machine Intelligence},
  author    = {Tan, Sinan and Ge, Mengmeng and Guo, Di and Liu, Huaping and Sun, Fuchun},
  month     = oct,
  year      = {2023},
  pages     = {11948--11960}
}

@article{garg_semantics_2020,
  title      = {Semantics for {Robotic} {Mapping}, {Perception} and {Interaction}: {A} {Survey}},
  volume     = {8},
  issn       = {1935-8253, 1935-8261},
  shorttitle = {Semantics for {Robotic} {Mapping}, {Perception} and {Interaction}},
  doi        = {10.1561/2300000059},
  language   = {en},
  number     = {1–2},
  urldate    = {2026-01-10},
  journal    = {Foundations and Trends® in Robotics},
  author     = {Garg, Sourav and Sünderhauf, Niko and Dayoub, Feras and Morrison, Douglas and Cosgun, Akansel and Carneiro, Gustavo and Wu, Qi and Chin, Tat-Jun and Reid, Ian and Gould, Stephen and Corke, Peter and Milford, Michael},
  year       = {2020},
  pages      = {1--224}
}

@article{loo_open_2025,
  title    = {Open scene graphs for open-world object-goal navigation},
  issn     = {0278-3649, 1741-3176},
  doi      = {10.1177/02783649251369549},
  language = {en},
  urldate  = {2026-01-11},
  journal  = {The International Journal of Robotics Research},
  author   = {Loo, Joel and Wu, Zhanxin and Hsu, David},
  month    = oct,
  year     = {2025},
  pages    = {02783649251369549}
}

@article{williams_simple_1992,
  title    = {Simple statistical gradient-following algorithms for connectionist reinforcement learning},
  volume   = {8},
  issn     = {1573-0565},
  doi      = {10.1007/BF00992696},
  language = {en},
  number   = {3},
  urldate  = {2025-06-17},
  journal  = {Machine Learning},
  author   = {Williams, Ronald J.},
  month    = may,
  year     = {1992},
  pages    = {229--256}
}

@misc{schulman_proximal_2017,
  title     = {Proximal {Policy} {Optimization} {Algorithms}},
  copyright = {arXiv.org perpetual, non-exclusive license},
  doi       = {10.48550/ARXIV.1707.06347},
  urldate   = {2025-08-06},
  publisher = {arXiv},
  author    = {Schulman, John and Wolski, Filip and Dhariwal, Prafulla and Radford, Alec and Klimov, Oleg},
  year      = {2017}
}

@article{tavakoli_action_2018,
  title    = {Action {Branching} {Architectures} for {Deep} {Reinforcement} {Learning}},
  volume   = {32},
  issn     = {2374-3468, 2159-5399},
  doi      = {10.1609/aaai.v32i1.11798},
  language = {en},
  number   = {1},
  urldate  = {2026-01-28},
  journal  = {Proceedings of the AAAI Conference on Artificial Intelligence},
  author   = {Tavakoli, Arash and Pardo, Fabio and Kormushev, Petar},
  month    = apr,
  year     = {2018}
}

@misc{engstrom_implementation_2020,
  title      = {Implementation {Matters} in {Deep} {Policy} {Gradients}: {A} {Case} {Study} on {PPO} and {TRPO}},
  shorttitle = {Implementation {Matters} in {Deep} {Policy} {Gradients}},
  doi        = {10.48550/arXiv.2005.12729},
  language   = {en},
  urldate    = {2026-01-14},
  publisher  = {arXiv},
  author     = {Engstrom, Logan and Ilyas, Andrew and Santurkar, Shibani and Tsipras, Dimitris and Janoos, Firdaus and Rudolph, Larry and Madry, Aleksander},
  month      = may,
  year       = {2020}
}

@misc{schulman_high-dimensional_2015,
  title     = {High-{Dimensional} {Continuous} {Control} {Using} {Generalized} {Advantage} {Estimation}},
  doi       = {10.48550/arXiv.1506.02438},
  urldate   = {2026-01-14},
  publisher = {arXiv},
  author    = {Schulman, John and Moritz, Philipp and Levine, Sergey and Jordan, Michael and Abbeel, Pieter},
  month     = oct,
  year      = {2015}
}

@article{hochreiter_long_1997,
  title    = {Long {Short}-{Term} {Memory}},
  volume   = {9},
  issn     = {0899-7667, 1530-888X},
  doi      = {10.1162/neco.1997.9.8.1735},
  language = {en},
  number   = {8},
  urldate  = {2026-01-14},
  journal  = {Neural Computation},
  author   = {Hochreiter, Sepp and Schmidhuber, Jürgen},
  month    = nov,
  year     = {1997},
  pages    = {1735--1780}
}

@article{ai2thor,
  author  = {Eric Kolve and Roozbeh Mottaghi and Winson Han and
             Eli VanderBilt and Luca Weihs and Alvaro Herrasti and
             Daniel Gordon and Yuke Zhu and Abhinav Gupta and
             Ali Farhadi},
  title   = {{AI2-THOR: An Interactive 3D Environment for Visual AI}},
  journal = {arXiv},
  year    = {2017}
}

@misc{kingma_adam_2017,
  title      = {Adam: {A} {Method} for {Stochastic} {Optimization}},
  shorttitle = {Adam},
  doi        = {10.48550/arXiv.1412.6980},
  urldate    = {2025-07-31},
  publisher  = {arXiv},
  author     = {Kingma, Diederik P. and Ba, Jimmy},
  month      = jan,
  year       = {2014}
}

@misc{akiba_optuna_2019,
  title      = {Optuna: {A} {Next}-generation {Hyperparameter} {Optimization} {Framework}},
  copyright  = {arXiv.org perpetual, non-exclusive license},
  shorttitle = {Optuna},
  doi        = {10.48550/ARXIV.1907.10902},
  urldate    = {2025-08-06},
  publisher  = {arXiv},
  author     = {Akiba, Takuya and Sano, Shotaro and Yanase, Toshihiko and Ohta, Takeru and Koyama, Masanori},
  year       = {2019}
}

@article{Chang_2023,
  title     = {A Comprehensive Survey of Scene Graphs: Generation and Application},
  volume    = {45},
  issn      = {1939-3539},
  doi       = {10.1109/tpami.2021.3137605},
  number    = {1},
  journal   = {IEEE Transactions on Pattern Analysis and Machine Intelligence},
  publisher = {Institute of Electrical and Electronics Engineers (IEEE)},
  author    = {Chang, Xiaojun and Ren, Pengzhen and Xu, Pengfei and Li, Zhihui and Chen, Xiaojiang and Hauptmann, Alex},
  year      = {2023},
  month     = jan,
  pages     = {1–26}
}

@article{embodied_ai_Duan,
  title     = {A Survey of Embodied AI: From Simulators to Research Tasks},
  author    = {Duan, Jiafei and Yu, Samson and Tan, Hui Li and Zhu, Hongyuan and Tan, Cheston},
  journal   = {IEEE Transactions on Emerging Topics in Computational Intelligence},
  volume    = {6},
  number    = {2},
  pages     = {230--244},
  year      = {2022},
  publisher = {IEEE}
}

@article{oc1_Tomforde,
  author  = {Tomforde, Sven and Sick, Bernhard and M{\"u}ller-Schloer, Christian},
  title   = {Organic Computing in the Spotlight},
  journal = {CoRR},
  volume  = {abs/1701.08125},
  year    = {2017},
  url     = {http://arxiv.org/abs/1701.08125}
}

@book{oc2_Mueller,
  title     = {Organic Computing-Technical Systems for Survival in the Real World},
  author    = {M{\"u}ller-Schloer, Christian and Tomforde, Sven},
  year      = {2017},
  publisher = {Springer}
}

@book{oc3_Mueller,
  title     = {Organic computing—a paradigm shift for complex systems},
  author    = {M{\"u}ller-Schloer, Christian and Schmeck, Hartmut and Ungerer, Theo},
  year      = {2011},
  publisher = {Springer Science \& Business Media}
}

@article{cur_lear_Narvekar,
  title   = {Curriculum learning for reinforcement learning domains: A framework and survey},
  author  = {Narvekar, Sanmit and Peng, Bei and Leonetti, Matteo and Sinapov, Jivko and Taylor, Matthew E and Stone, Peter},
  journal = {Journal of Machine Learning Research},
  volume  = {21},
  number  = {181},
  pages   = {1--50},
  year    = {2020}
}

@inproceedings{auto_explo_yamauchi,
  title        = {A frontier-based approach for autonomous exploration},
  author       = {Yamauchi, Brian},
  booktitle    = {Proceedings 1997 IEEE International Symposium on Computational Intelligence in Robotics and Automation CIRA'97.'Towards New Computational Principles for Robotics and Automation'},
  pages        = {146--151},
  year         = {1997},
  organization = {IEEE}
}

@inproceedings{NBV_bircher,
  title        = {Receding horizon "next-best-view" planner for 3d exploration},
  author       = {Bircher, Andreas and Kamel, Mina and Alexis, Kostas and Oleynikova, Helen and Siegwart, Roland},
  booktitle    = {2016 IEEE international conference on robotics and automation (ICRA)},
  pages        = {1462--1468},
  year         = {2016},
  organization = {IEEE}
}

@article{act_neur_slam_chaplot,
  title   = {Learning to explore using active neural slam},
  author  = {Chaplot, Devendra Singh and Gandhi, Dhiraj and Gupta, Saurabh and Gupta, Abhinav and Salakhutdinov, Ruslan},
  journal = {arXiv preprint arXiv:2004.05155},
  year    = {2020}
}

@article{dd_ppo_wijmans,
  title   = {DD-PPO: Learning near-perfect pointgoal navigators from 2.5 billion frames},
  author  = {Wijmans, Erik and Kadian, Abhishek and Morcos, Ari and Lee, Stefan and Essa, Irfan and Parikh, Devi and Savva, Manolis and Batra, Dhruv},
  journal = {arXiv preprint arXiv:1911.00357},
  year    = {2019}
}

@inproceedings{sem_mapnet_cartillier,
  title     = {Semantic mapnet: Building allocentric semantic maps and representations from egocentric views},
  author    = {Cartillier, Vincent and Ren, Zhile and Jain, Neha and Lee, Stefan and Essa, Irfan and Batra, Dhruv},
  booktitle = {Proceedings of the AAAI Conference on Artificial Intelligence},
  volume    = {35},
  pages     = {964--972},
  year      = {2021}
}

\appendix
\section{Appendix}
\subsection{Figures}
\label{app:figures}

\begin{figure}[H]
    \centering
    \setlength{\textwidth}{0.9\textwidth}
    \begin{minipage}{0.32\textwidth}
        \centering
        \caption*{\textit{FloorPlan 28}}
        \includegraphics[width=\linewidth]{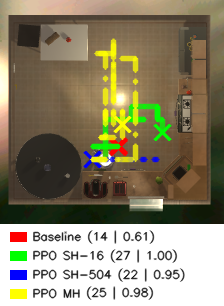}
    \end{minipage}
    \begin{minipage}{0.32\textwidth}
        \centering
        \caption*{\textit{FloorPlan 29}}
        \includegraphics[width=\linewidth]{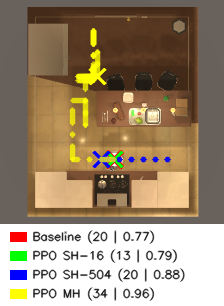}
    \end{minipage}
    \begin{minipage}{0.32\textwidth}
        \centering
        \caption*{\textit{FloorPlan 30}}

        \includegraphics[width=\linewidth]{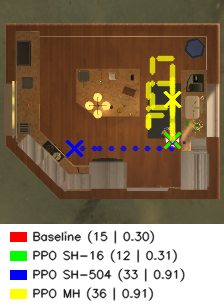}
    \end{minipage}

    \vspace{0.5em}

    \begin{minipage}{0.32\textwidth}
        \centering
        \includegraphics[width=\linewidth]{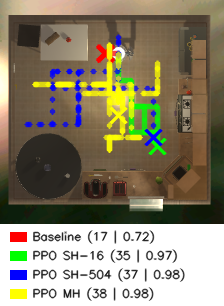}
    \end{minipage}
    \begin{minipage}{0.32\textwidth}
        \centering
        \includegraphics[width=\linewidth]{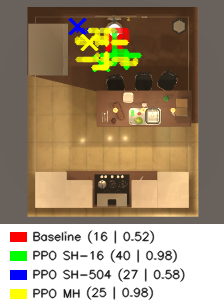}
    \end{minipage}
    \begin{minipage}{0.32\textwidth}
        \centering
        \includegraphics[width=\linewidth]{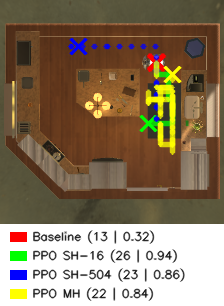}
    \end{minipage}

    \vspace{0.5em}

    \begin{minipage}{0.32\textwidth}
        \centering
        \includegraphics[width=\linewidth]{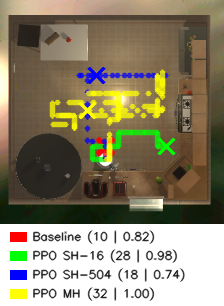}
    \end{minipage}
    \begin{minipage}{0.32\textwidth}
        \centering
        \includegraphics[width=\linewidth]{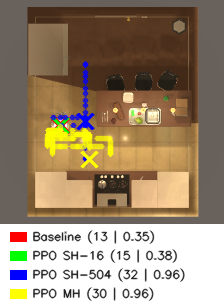}
    \end{minipage}
    \begin{minipage}{0.32\textwidth}
        \centering
        \includegraphics[width=\linewidth]{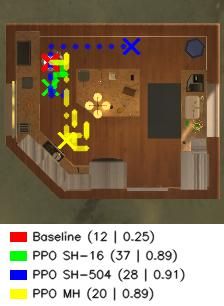}
    \end{minipage}

    \caption{Representative trajectories on held-out scenes (FloorPlans 28--30).
        Numbers in parentheses in the legends report \texttt{(steps | score)}.
        All plotted scenarios except the Baseline are with depth input.}
    \label{fig:qual-trajectories-app}
\end{figure}

\subsection{Final Held-Out Metrics}
\label{app:final-metrics}
The metrics in Table~\ref{tab:final-metrics} are computed via a post-hoc evaluation on the held-out scenes (FloorPlans~28--30) using the final training checkpoint of each run (one checkpoint per seed), averaged over episodes and scenes per seed, and then aggregated across seeds.

\begin{table}[ht]
    \centering
    \caption{Final held-out performance on FloorPlans~28--30 using the final training checkpoint of each run (means across seeds; Node Recall reported as mean $\pm$ std).\\
        \textit{Node Recall}: fraction of scene graph nodes recalled at episode end.\\
        \textit{Episodic Return}: cumulative reward under the composite reward function.\\
        \textit{Episode Length}: number of steps taken until episode termination.\\
        \textit{Move Success Rate}: fraction of executed translation actions that succeed without collision.\\
        \textit{Path Length}: total path length traversed during the episode.
        \\ \textbf{ $\uparrow$} indicates that higher values are better.}
    \label{tab:final-metrics}
    \begin{tabular}{lccccc}
        \toprule
        \textbf{Scenario}    & \shortstack[c]{\textbf{Node}                                                        \\\textbf{Recall}\\ \textbf{ $\uparrow$}} & \shortstack[c]{\textbf{Episodic}\\\textbf{Return} \\ \textbf{ $\uparrow$}} & \shortstack[c]{\textbf{Episode}\\\textbf{Length}} & \shortstack[c]{\textbf{Move} \\ \textbf{Success}\\\textbf{Rate}\\ \textbf{ $\uparrow$}} & \shortstack[c]{\textbf{Path}\\\textbf{Length} \\ \textbf{ $\uparrow$}} \\
        \midrule
        Baseline (R+SH16+IL) & 0.48 $\pm$ 0.22              & 0.69          & 14.0 & 0.14          & 0.1           \\
        S0 (R+SH16+IL+D)     & 0.50 $\pm$ 0.24              & 0.69          & 15.4 & 0.14          & 0.2           \\
        S1 (R+SH16+D)        & 0.50 $\pm$ 0.23              & 0.69          & 16.2 & 0.15          & 0.2           \\
        S2 (PPO+SH16)        & 0.58 $\pm$ 0.29              & 0.84          & 21.0 & 0.71          & 1.6           \\
        S3 (PPO+SH16+D)      & 0.58 $\pm$ 0.29              & 0.86          & 20.8 & \textbf{0.77} & 2.2           \\
        S4 (PPO+SH504+D)     & 0.92 $\pm$ 0.14              & 0.89          & 32.7 & 0.42          & 7.9           \\
        S5 (PPO+SH504+D+CL)  & 0.86 $\pm$ 0.19              & 0.79          & 28.3 & 0.30          & 4.2           \\
        S6 (PPO+MH504)       & \textbf{0.93} $\pm$ 0.14     & 1.07          & 27.7 & 0.56          & 10.8          \\
        S7 (PPO+MH504+D)     & 0.91 $\pm$ 0.15              & 1.04          & 26.0 & 0.56          & 10.1          \\
        S8 (PPO+MH504+CL)    & \textbf{0.93} $\pm$ 0.14     & 1.09          & 27.8 & 0.57          & 12.2          \\
        S9 (PPO+MH504+D+CL)  & \textbf{0.93} $\pm$ 0.15     & \textbf{1.11} & 29.3 & 0.58          & \textbf{12.9} \\
        \bottomrule
    \end{tabular}
\end{table}

\subsection{Scenario Overview}
\label{app:scenarios}
To isolate the contribution of individual design choices, we evaluate a set of ablation scenarios that progressively modify the learning setup. The scenarios differ in the RL algorithm, the presence of imitation learning (IL) pretraining, the action parameterisation (single-head vs.\ multi-head), the use of depth input with a collision-prediction auxiliary head, and whether curriculum learning (CL) with staged action-space expansion is enabled.

\begin{table}[H]
    \centering
    \caption{Overview of ablation scenarios.\\
        \textit{IL}: imitation learning pretraining.\\
        \textit{Actions}: single-head parameterisation (either compact 16 actions or enlarged 504 actions) versus factorised multi-head parameterisation (Rotation angle, Length, Stop) at 504 resolution.\\
        \textit{Depth w/ Coll.-Aux}: depth input with an auxiliary head for collision prediction.\\
        \textit{CL}: curriculum learning with staged action-space expansion.}
    \label{tab:scenarios}
    \small
    \setlength{\tabcolsep}{6pt}
    \renewcommand{\arraystretch}{1.15}
    \begin{tabular}{lccccc}
        \toprule
        \textbf{Scenario} & \textbf{RL-Algorithm} & \textbf{IL} & \textbf{Action Variant} & \textbf{\shortstack{Depth w/              \\ Coll.-Aux}} & \textbf{CL} \\
        \midrule
        Baseline          & REINFORCE             & \checkmark  & Single-Head (16)        & --                           & --         \\
        S0                & REINFORCE             & \checkmark  & Single-Head (16)        & \checkmark                   & --         \\
        S1                & REINFORCE             & --          & Single-Head (16)        & \checkmark                   & --         \\
        S2                & PPO                   & --          & Single-Head (16)        & --                           & --         \\
        S3                & PPO                   & --          & Single-Head (16)        & \checkmark                   & --         \\
        S4                & PPO                   & --          & Single-Head (504)       & \checkmark                   & --         \\
        S5                & PPO                   & --          & Single-Head (504)       & \checkmark                   & \checkmark \\
        S6                & PPO                   & --          & Multi-Head (504)        & --                           & --         \\
        S7                & PPO                   & --          & Multi-Head (504)        & \checkmark                   & --         \\
        S8                & PPO                   & --          & Multi-Head (504)        & --                           & \checkmark \\
        S9                & PPO                   & --          & Multi-Head (504)        & \checkmark                   & \checkmark \\
        \bottomrule
    \end{tabular}
\end{table}

\subsection{Imitation Learning Dataset Generation}
\label{app:il-dataset}

\paragraph{Overview and stored format.}
We generate an offline dataset of expert trajectories in AI2-THOR for imitation learning pretraining of the baseline.
Each trajectory is stored as a sequence of samples $\{(x_t, a_t)\}_{t=0}^{T-1}$, where $x_t$ denotes the state representation used during training (RGB, optional depth, local/global scene-graph information, and auxiliary signals such as the previous action), and $a_t$ is a discrete navigation action from the 16-action \textit{single-head} space (8 rotation angles $\times$ 2 lengths, with \textsc{Stop} at rotation angle $=0$ and length $=0$).
The rotation angles are multiples of $45^\circ$ and the lengths are $\{0\,\text{m}, 0.3\,\text{m}\}$.
For each sample we additionally store metadata (scene identifier, start pose, step index).
Every stored trajectory terminates with an explicit \textsc{Stop} action.

\paragraph{Start state selection.}
For each training scene, we sample multiple start states from the set of reachable positions.
A start state is accepted only if movement is possible from that position.

\paragraph{Viewpoint targets and tour ordering.}
As target structure, we use a precomputed \textit{viewpoint-to-objects} mapping that associates each viewpoint with locally visible objects and their corresponding visibility values.
We greedily select a set of viewpoints that maximises the aggregated visibility progress over objects that are not yet sufficiently covered (visibility aggregation follows the multiplicative update rule explained in Appendix~\ref{app:soft-visibility}).
The selected viewpoints are then ordered into a short tour by solving a travelling salesman problem (TSP) heuristic using Ant-Colony Optimization over the viewpoint positions.
The resulting tour is rotated such that it starts at the viewpoint closest to the sampled start state.

\paragraph{Expert action selection (short-horizon planning).}
The expert follows the current target viewpoint of the tour and selects navigation actions via short-horizon planning.
To this end, multiple admissible action candidates are forward-simulated in a separate simulator instance and scored using a heuristic objective that combines
(i) visibility gain and first-time crossing of $\tau$ (``discovery''),
(ii) a bonus for entering previously unvisited grid viewpoints, and
(iii) progress towards the target viewpoint.
Candidates whose execution fails (e.g., due to blocking) are discarded.
In addition, we exclude candidates that perform a rotation but would leave the agent without any valid forward translation afterward (two-step validation consistent with the \textit{move-first} action semantics).

\paragraph{Termination and early stopping.}
We retain only trajectories that terminate successfully (via \textsc{Stop}) or achieve near-complete graph coverage.
In addition to a maximum step budget $T_{\max}$ of 80 steps, we apply early stopping: once at least $95\%$ of ground-truth objects are considered discovered (visibility $\ge \tau$) and no further progress is observed over several steps, we remove stagnating tail steps and append a final \textsc{Stop} action.
Expert trajectories are generated offline with a higher step budget ($T_{\max} = 80$) to increase coverage, while online RL episodes remain constrained to 40 steps.
\subsection{Stagnation Heuristic}
\label{app:stagnation-formulas}
To help the policy decide when further navigation yields diminishing returns, we provide a compact \emph{stagnation signal} that measures how strongly the global semantic scene graph (GSSG) changes over time.
Intuitively, if the GSSG embedding barely changes for multiple steps, the agent is likely revisiting already-covered areas, while strong changes indicate ongoing discovery.
We encode this signal as an additional state feature used by the navigation policy.

All parameters used are reported in Table~\ref{tab:stagnation_hyperparameters}.
We define the change magnitude of the global scene-graph representation as the normalised $\ell_2$ difference
\begin{equation}
    \Delta_t
    =
    \frac{\lVert \mathbf{g}_t - \mathbf{g}_{t-1} \rVert_2}{\sqrt{d_{\text{sg}}}},
    \label{eq:stag-delta}
\end{equation}
where $\mathbf{g}_t \in \mathbb{R}^{d_{\text{sg}}}$ denotes the GSSG embedding and $d_{\text{sg}}$ is its dimensionality.
If no previous embedding is available (sequence start), we set $\Delta_1 := 1$.

\begin{table}[htbp]
    \centering
    \caption{Hyperparameters for stagnation detection.}
    \label{tab:stagnation_hyperparameters}
    \small
    \setlength{\tabcolsep}{6pt}
    \renewcommand{\arraystretch}{1.15}
    \begin{tabular}{ll}
        \toprule
        \textbf{Parameter}                               & \textbf{Value} \\
        \midrule
        EMA smoothing $\alpha$                           & 0.7            \\
        Embedding change threshold $\tau_{\text{disc}}$  & 0.1            \\
        Scene-graph dimension $d_{\text{sg}}$            & 512            \\
        Divisor $k$                                      & 10             \\
        Stagnation embedding dimension $d_{\text{stag}}$ & 32             \\
        \bottomrule
    \end{tabular}
\end{table}

For smoothing, we use an exponential moving average (EMA):
\begin{equation}
    \bar{\Delta}_t = \alpha \bar{\Delta}_{t-1} + (1-\alpha)\Delta_t,
    \label{eq:stag-ema}
\end{equation}
initialised with $\bar{\Delta}_1 := \Delta_1$.
An \emph{embedding change} is detected when $\Delta_t > \tau_{\text{disc}}$, indicating a sufficiently large change in the global scene-graph embedding.
We define the indicator
\begin{equation}
    d_t = \mathbb{I}\!\left[\Delta_t > \tau_{\text{disc}}\right].
    \label{eq:stag-discovery-ind}
\end{equation}
To quantify how long the agent has been navigating without producing new semantic information, we track the number of steps since the last embedding change.
The counter \texttt{steps\_\allowbreak since\_\allowbreak embedding\_\allowbreak change} is denoted by $n_t$ and updated via masking:
\begin{equation}
    n_t = (1-d_t)\,n_{t-1} + 1,
    \label{eq:stag-steps-mask}
\end{equation}
with initialisation $n_1 := 1$.

The stagnation signal is then defined as
\begin{equation}
    s_t = \tanh\!\left(\frac{n_t}{k}\right),
    \label{eq:stag-tanh}
\end{equation}
where $k$ is chosen empirically (in our experiments $k=10$).

As input to the stagnation encoder, we use the concatenation of the three scalar quantities
\begin{equation}
    \mathbf{x}^{\text{stag}}_t = \left[\Delta_t,\; \bar{\Delta}_t,\; s_t\right] \in \mathbb{R}^3.
    \label{eq:stag-concat}
\end{equation}
The stagnation embedding $\mathbf{e}^{\text{stag}}_t \in \mathbb{R}^{d_{\text{stag}}}$ is then produced by a two-layer MLP:
\begin{align}
    \mathbf{h}_t               & = \text{ReLU}\!\left(\mathbf{W}_1 \mathbf{x}^{\text{stag}}_t + \mathbf{b}_1\right), \quad \mathbf{W}_1 \in \mathbb{R}^{64 \times 3},
    \label{eq:stag-mlp-1}                                                                                                                                             \\
    \mathbf{e}^{\text{stag}}_t & = \mathbf{W}_2 \mathbf{h}_t + \mathbf{b}_2, \quad \mathbf{W}_2 \in \mathbb{R}^{d_{\text{stag}}\times 64}.
    \label{eq:stag-mlp-2}
\end{align}

\subsection{Soft Visibility and Global Visibility Update}
\label{app:soft-visibility}
We encode object visibility as a continuous score that reflects how reliably an object would be observable from the current viewpoint based on distance and object size.
Let $d$ denote the agent--object distance and $s_{\max}$ a size proxy (e.g., maximum 2D/3D extent available from simulator metadata).
We compute a soft-visibility score via a sigmoid with a size-dependent centre:
\begin{equation}
    V_{\text{soft}}(d, s_{\max})
    =
    \frac{1}{1+\exp\!\left(\omega\left(d - c(s_{\max})\right)\right)},
    \label{eq:softvis}
\end{equation}
where
\begin{equation}
    c(s_{\max})
    =
    c_{\text{base}} + k_{\text{size}}\left(s_{\max}-s_{\text{ref}}\right).
    \label{}
\end{equation}
Here, $c_{\text{base}}$ is the baseline visibility range (in metres) for an object of reference size $s_{\text{ref}}$,
and $k_{\text{size}}$ determines how strongly the sigmoid centre is shifted along the distance axis with object size.
The parameter $\omega$ determines the steepness of the distance-to-visibility transition.
We use empirically selected parameters $\omega=1.0$, $c_{\text{base}}=3.5$, $k_{\text{size}}=1.5$, and $s_{\text{ref}}=0.5$ that yield a visibility measure consistent with common-sense expectations: larger objects remain visible at larger distances, whereas small objects require closer viewpoints to reach high visibility.

\paragraph{Global aggregation.}
The GSSG is updated by aggregating local observations over time.
Let $V^{G}_{t-1}(o)\in[0,1]$ denote the global visibility of object $o$ at time $t-1$, and let $V^{L}_{t}(o)=V_{\text{soft}}(d_t(o),s_{\max}(o))$ denote the local soft visibility at time $t$.
We update the global visibility with a monotonic, saturating rule: each new observation increases $V^{G}$, while the estimate remains bounded in $[0,1]$ and exhibits diminishing returns as $V^{G}$ approaches $1$:
\begin{equation}
    V^{G}_{t}(o)
    =
    1 - \left(1 - V^{G}_{t-1}(o)\right)\left(1 - V^{L}_{t}(o)\right).
    \label{eq:softvis-agg}
\end{equation}
A node is counted as discovered if $V^{G}_{t}(o)\ge \tau$, where $\tau=0.8$ is a discovery threshold.

\paragraph{Redundancy control.}
To reduce redundant updates, we skip the global update for an object if it has already been observed from the same agent position (i.e., repeated rotations at an identical location do not increase visibility).
This encourages spatial coverage rather than repeated re-observation from identical viewpoints.

\paragraph{Best-case assumption.}
Both LSSG extraction and GSSG aggregation are derived from simulator metadata with perfect data association via object identifiers.
Consequently, soft-visibility models only \emph{visibility} and not perception failures; the resulting completeness metrics should be interpreted as an idealised best-case reference.

\subsection{Relation Extraction from Simulator Metadata}
\label{app:relation-extraction}

To construct semantic relations in the Local Semantic Scene Graph (LSSG), we derive edges deterministically from AI2-THOR metadata using rule-based geometric and type-specific heuristics, following the oracle-style graph construction protocol of Li et al. in their paper “Embodied Semantic Scene Graph Generation”.
All relations are computed per time step from the set of currently visible objects and their metadata fields (notably \code{objectType}, \code{objectId}, \code{parentReceptacles}, and the axis-aligned bounding box centre/size).

We generate directed relation pairs with inverse labels where applicable, yielding a multi-relational graph.
The relation set comprises:

\noindent(i) \emph{support and placement} relations between movable objects and supporting surfaces, inferred from bounding-box overlap in the horizontal plane and small vertical offsets (\code{supported_by}/\code{supports}, and \code{on}/\code{has_on_top} with optional subtypes such as \code{standing_on}, \code{sitting_on}, \code{lying_on});

\noindent(ii) \emph{hanging} relations for wall-mounted objects, mapped from object categories to relation labels (e.g., \code{hanging_on}, \code{pasting_on}, \code{fixed_on});

\noindent(iii) \emph{relative-position} relations between objects that share the same receptacle, based on 3D proximity and vertical ordering (\code{close_by}, \code{above}/\code{below});

\noindent(iv) \emph{connectivity} relations for adjacent furniture instances of the same type (\mbox{\code{connects_to}}) and

\noindent(v) \emph{attachment/component} relations for small parts that are spatially close to a corresponding main object (\code{attach_on}/\code{has_attachment}, \code{part_of}/\code{has_part}).

Since relations are derived deterministically from simulator metadata, they are noise-free in our setting and thus represent an idealised upper bound compared to pipelines that infer relations from learned perception and data association.
\subsection{Event-based Reward Terms}
\label{app:reward-terms}

In addition to the dense potential-based shaping term, we include sparse event-based reward components that encourage translation (especially collision-free translation), reward viewpoint discoveries, and provide an explicit stopping incentive once exploration has stagnated.
Coefficients were calibrated on the training scenes to obtain stable learning signals and comparable reward magnitudes, then kept fixed across all ablation scenarios to isolate the effects of algorithmic and action-space changes; held-out evaluation scenes are used only for reporting.

\begin{align}
    r_t^{\text{coll}}
               & =
    -0.02 \cdot \mathbb{I}\!\left[\texttt{move\_failed} \,\lor\, \texttt{actual\_dist} < \texttt{target\_dist} - 0.05\right],
    \label{eq:reward-coll}                                                                                                    \\
    r_t^{\text{move}}
               & =
    0.005 \cdot \mathbb{I}\!\left[\texttt{actual\_dist} > 0.05\right],
    \label{eq:reward-move}                                                                                                    \\
    r^{expl}_t & = 0.01 \cdot \mathbb{I}\!\left[\Delta |\mathcal{O}_{seen}| > 0\right],
    \label{eq:reward-expl}                                                                                                    \\
    r^{stop}_t & = b_{stop} \cdot \mathbb{I}\!\left[a_t=\mathrm{Stop} \wedge \texttt{steps\_since\_exploration} \ge K\right],
    \label{eq:reward-stop}
\end{align}
Here, \code{target_dist} denotes the translation distance requested by the action,
\code{actual_dist} is the distance actually traveled in the environment,
$\mathcal{O}_{seen}$ denotes the set of object identifiers that have been observed at least once so far
(i.e., the set of keys in the viewpoint-tracking map),
\code{move_failed} indicates a failed motion step (e.g., due to collision or blocking),
and \texttt{steps\_\allowbreak since\_\allowbreak exploration} denotes the number of steps since the last first-time object observation event.
\subsection{Adaptive Curriculum for Large Discrete Action Spaces}
\label{app:curriculum}
All relevant hyperparameters are listed in Table~\ref{tab:curriculum_hyperparameters}.
We use a staged curriculum to ease optimisation in large discrete action spaces.
The curriculum is applied to both the enlarged single-head action space (504 actions) and the factorised multi-head parameterisation (Rotation angle, Length, \textsc{Stop}) at the same 504-action resolution.
In each stage, only a restricted subset of actions (single-head) or action-factor combinations (multi-head) is admissible, and invalid choices are masked by setting their logits to $-\infty$.

\begin{table}[htbp]
    \centering
    \caption{Curriculum learning hyperparameters for large action spaces.}
    \label{tab:curriculum_hyperparameters}
    \small
    \setlength{\tabcolsep}{6pt}
    \renewcommand{\arraystretch}{1.15}
    \begin{tabular}{ll}
        \toprule
        \textbf{Parameter}                     & \textbf{Value} \\
        \midrule
        Min.\ stage blocks                     & 50             \\
        Plateau evaluation window $W$ (blocks) & 50             \\
        Recent window $R$ (blocks)             & 10             \\
        Plateau threshold                      & 0.02           \\
        Forced-promotion backstop (blocks)     & 250            \\
        Entropy boost factor                   & 2.0            \\
        Entropy boost duration (blocks)        & 20             \\
        EMA smoothing $\alpha$                 & 0.2            \\
        \bottomrule
    \end{tabular}
\end{table}

\paragraph{Stages.}
We progressively increase action-space resolution over four stages (Table~\ref{tab:curriculum_stages}).
In the multi-head parameterisation, stages restrict admissible rotation-angle and length indices.
In the single-head parameterisation, stages restrict admissible atomic actions to the corresponding Cartesian product.
The \textsc{Stop} decision remains available throughout.
Stage~3 uses a near-complete 0.1\,m grid; we exclude 1.9\,m to keep the stage at 20 lengths while ensuring the length set grows monotonically and contains all Stage~2 values.
\begin{table}[htbp]
    \centering
    \caption{Curriculum stages for large action spaces (\textsc{Stop} available throughout). Lengths in metres.}
    \label{tab:curriculum_stages}
    \small
    \setlength{\tabcolsep}{5pt}
    \renewcommand{\arraystretch}{1.15}
    \begin{tabular}{lcccll}
        \toprule
        \textbf{Stage} & \textbf{\#Angles} & \textbf{Angle step} & \textbf{\#Lengths} & \textbf{Total} & \textbf{Lengths}                        \\
        \midrule
        1              & 8                 & $45^\circ$          & 2                  & 16             & $\{0.0,0.3\}$                           \\
        2              & 8                 & $45^\circ$          & 6                  & 48             & $\{0.0,0.3,0.7,1.2,1.6,2.0\}$           \\
        3              & 8                 & $45^\circ$          & 20                 & 160            & $\{0.0,0.1,\dots,2.0\}\setminus\{1.9\}$ \\
        4              & 24                & $15^\circ$          & 21                 & 504            & $\{0.0,0.1,\dots,2.0\}$                 \\
        \bottomrule
    \end{tabular}
\end{table}

\paragraph{Plateau detection.}
To decide when to promote to the next stage, we track block-aggregated rollout statistics and smooth them via an exponential moving average (EMA) with smoothing factor $\alpha$ (Table~\ref{tab:curriculum_hyperparameters}) to reduce noise.
Specifically, we monitor (i) the primary task score (Node Recall), (ii) the mean episodic return, and (iii) the mean episode length (steps), since these three signals jointly capture coverage progress, optimisation progress under the shaped reward, and termination behaviour.
Plateau decisions are evaluated every $W$ blocks, where $W$ is the \emph{plateau evaluation window}.
For each metric, we compare the mean value over the most recent $R$ blocks (the \emph{recent window}) against the mean value over the preceding $W\!-\!R$ blocks (the \emph{earlier window}) on the smoothed curves.
A promotion is triggered if the relative improvement between recent and earlier window is below the plateau threshold for all monitored metrics.

\paragraph{Backstop and minimum stage duration.}
To prevent the curriculum from stalling indefinitely, we (i) enforce a minimum number of blocks spent in each stage and (ii) apply a backstop that promotes after a fixed block budget even if plateau conditions are not met.

\paragraph{Entropy ramp after promotion.}
After promotion, the entropy coefficient is temporarily increased and then linearly annealed back to its default value over a fixed number of blocks.
This encourages the policy to explore the newly available actions rather than remaining confined to the previous stage’s behaviour.

\subsection{Collision Auxiliary Head and Loss}
\label{app:collision-aux}
The collision head is trained as a binary classifier and estimates, per time step, the probability that the last \emph{translation} step could not be completed due to blockage or collision.
As supervision, we use the environment's binary motion feedback (success vs.\ failure) and define a label $y_t \in \{0,1\}$ (1: collision/blockage, 0: successful motion).

Importantly, we supervise the collision head \emph{only} on actions with non-zero translation length.
Pure rotation steps (i.e., length $=0$) are masked out and do not contribute to the loss.
If a mini-batch contains no translation actions, the collision loss is set to zero.

We optimise binary cross entropy with logits on the head logits $z_t$:
\[
    \mathcal{L}_{\text{collision}}
    =
    \frac{1}{\sum_{t=1}^{T} m_t}\sum_{t=1}^{T} m_t \cdot \mathrm{BCEWithLogits}(z_t, y_t),
\]
where $m_t \in \{0,1\}$ is a mask indicating whether step $t$ is a translation action (1) or a rotation-only action (0).
\subsection{Optuna Objective}
\label{app:optuna-objective}
To select hyperparameters in a reproducible and task-aligned manner, we optimise a single scalar objective that prioritises scene-graph completeness while applying light secondary penalties for collisions, truncations, and overly long episodes on a recent rollout window.
\begin{equation}
    \label{eq:optuna-objective}
    J
    =
    \bar{s}
    -
    0.20\,\bar{c}
    -
    0.10\,\bar{\tau}
    -
    0.01\,\max\!\left(0,\,\bar{n}-25\right),
\end{equation}
\noindent
where $\bar{s}$ is the mean episode score (Node Recall) within the considered rollout window,
$\bar{n}$ is the mean episode length (in steps),
$\bar{\tau}$ is the truncation rate (fraction of episodes terminated by the step limit),
and $\bar{c}$ is the mean collision rate in the same window.
The window spans the most recent $\approx 50$ training blocks.
For numerical stability, we clip the objective from below at $-100$.

\paragraph{Optimisation budget and pruning.}
Hyperparameters are optimised with Optuna using a TPE sampler ($5$ startup trials).
Each trial trains a policy for a fixed number of blocks, depending on the scenario, to account for different learning speeds (Table~\ref{tab:optuna-budget}).
Early stopping is implemented via Optuna pruning: we report $J$ at block resolution and prune underperforming trials after a warmup phase.
For REINFORCE+IL, PPO with compact actions and multi-head large action space, we use a MedianPruner; for the slower-learning single-head large action space we use a less aggressive PercentilePruner (bottom $25\%$).
In all cases, pruning is checked every block and only enabled after the respective warmup; pruners are additionally activated only after a small number of completed trials (startup trials), following Optuna defaults.
All Optuna tuning runs are performed with depth and without curriculum learning.

\begin{table}[t]
    \centering
    \caption{Per-trial training budget and pruning configuration used during hyperparameter optimisation. PPO entries listed as “SH16 / SH504 / MH” correspond to the compact single-head action space (16 actions),
        the full single-head action space (504 actions), and the multi-head factorised action space (24 rotation angles, 21 lengths, \textsc{Stop}), respectively.}
    \label{tab:optuna-budget}
    \begin{tabular}{lcccc}
        \toprule
        Scenario     & Blocks / trial & Warmup (blocks) & Pruner            & Startup trials \\
        \midrule
        REINFORCE+IL & 400            & 200             & Median            & 2              \\
        PPO SH16     & 400            & 200             & Median            & 2              \\
        PPO SH504    & 800            & 600             & Percentile (25\%) & 3              \\
        PPO MH       & 500            & 300             & Median            & 2              \\

        \bottomrule
    \end{tabular}
\end{table}

\begin{table}[ht]
    \centering
    \caption{Optuna hyperparameter search spaces.
        All parameters were optimised over categorical candidate sets.
        The optimisation target is the composite objective in Eq.~\ref{eq:optuna-objective}.}
    \label{tab:optuna-search-space}
    \small
    \setlength{\tabcolsep}{4pt}
    \renewcommand{\arraystretch}{1.15}
    \begin{tabular}{lll}
        \toprule
        \textbf{Parameter}           & \textbf{Search Space (categorical)}                                       & \textbf{Best}  \\
        \midrule
        \multicolumn{3}{l}{\textbf{REINFORCE (all setups)}}                                                                       \\
        $\alpha$                     & $\{1\text{e-}5,\,3\text{e-}5,\,5\text{e-}5,\,1\text{e-}4,\,5\text{e-}4\}$ & \textbf{5e-4}  \\
        $\gamma$                     & $\{0.90,\,0.95,\,0.97,\,0.99\}$                                           & \textbf{0.97}  \\
        $c_{\text{aux}}$ (collision) & $\{0.2,\,0.3,\,0.4\}$                                                     & \textbf{0.3}   \\
        $c_{\text{ent}}$ (entropy)   & $\{0.01,\,0.05,\,0.10,\,0.15\}$                                           & \textbf{0.05}  \\
        \addlinespace
        \multicolumn{3}{l}{\textbf{PPO (Single-Head, 16 actions)}}                                                                \\
        $\alpha$                     & $\{1\text{e-}5,\,3\text{e-}5,\,5\text{e-}5,\,1\text{e-}4,\,5\text{e-}4\}$ & \textbf{1e-4}  \\
        $\epsilon$ (clip)            & $\{0.10,\,0.15,\,0.20,\,0.25,\,0.30\}$                                    & \textbf{0.25}  \\
        $\lambda$ (GAE)              & $\{0.90,\,0.93,\,0.95,\,0.97,\,0.98\}$                                    & \textbf{0.90}  \\
        $c_{\text{aux}}$ (collision) & $\{0.2,\,0.3,\,0.4\}$                                                     & \textbf{0.2}   \\
        $c_{\text{ent}}$ (entropy)   & $\{0.01,\,0.05,\,0.10,\,0.15\}$                                           & \textbf{0.05}  \\
        \addlinespace
        \multicolumn{3}{l}{\textbf{PPO (Single-Head, 504 actions)}}                                                               \\
        $\alpha$                     & $\{1\text{e-}5,\,3\text{e-}5,\,5\text{e-}5,\,1\text{e-}4,\,5\text{e-}4\}$ & \textbf{1e-4}  \\
        $\epsilon$ (clip)            & $\{0.10,\,0.15,\,0.20,\,0.25,\,0.30\}$                                    & \textbf{0.15}  \\
        $\lambda$ (GAE)              & $\{0.90,\,0.93,\,0.95,\,0.97,\,0.98\}$                                    & \textbf{0.97}  \\
        $c_{\text{aux}}$ (collision) & $\{0.2,\,0.3,\,0.4\}$                                                     & \textbf{0.3}   \\
        $c_{\text{ent}}$ (entropy)   & $\{0.01,\,0.05,\,0.10,\,0.15\}$                                           & \textbf{0.05 } \\
        \addlinespace
        \multicolumn{3}{l}{\textbf{PPO (Multi-Head, 504 resolution)}}                                                             \\
        $\alpha$                     & $\{1\text{e-}5,\,3\text{e-}5,\,5\text{e-}5,\,1\text{e-}4,\,5\text{e-}4\}$ & \textbf{1e-4}  \\
        $\epsilon$ (clip)            & $\{0.10,\,0.15,\,0.20,\,0.25,\,0.30\}$                                    & \textbf{0.15}  \\
        $\lambda$ (GAE)              & $\{0.90,\,0.93,\,0.95,\,0.97,\,0.98\}$                                    & \textbf{0.97}  \\
        $c_{\text{aux}}$ (collision) & $\{0.2,\,0.3,\,0.4\}$                                                     & \textbf{0.3}   \\
        $c_{\text{ent}}$ (entropy)   & $\{0.01,\,0.05,\,0.10,\,0.15\}$                                           & \textbf{0.05}  \\
        \bottomrule
    \end{tabular}
\end{table}

\clearpage
\subsection{Further Tables}
\label{app:tables}
\begin{table}[H]
    \centering
    \caption{Feature dimensions and architecture specifications.}
    \label{tab:dimensions}
    \small
    \setlength{\tabcolsep}{4pt}
    \renewcommand{\arraystretch}{1}
    \begin{tabular}{lll}
        \toprule
        \textbf{\textit{Component}} & \textbf{Parameter}                  & \textbf{Value}                          \\
        \midrule
        \multicolumn{3}{l}{\textit{Input dimensions}}                                                               \\
                                    & RGB Image                           & 224$\times$224$\times$3                 \\
                                    & Depth Map (optional)                & 224$\times$224$\times$1                 \\
        \hline
        \multicolumn{3}{l}{\textit{Feature Extractors}}                                                             \\
                                    & RGB encoder (ResNet18)              & 512-dim                                 \\
                                    & Depth CNN layers                    & 3                                       \\
                                    & Depth encoder output                & 256-dim                                 \\
                                    & Action embedding                    & 64-dim                                  \\
                                    & Stagnation embedding                & 32-dim                                  \\
        \hline
        \multicolumn{3}{l}{\textit{Graph encoder (HGT)}}                                                            \\
                                    & LSSG embedding                      & 512-dim                                 \\
                                    & GSSG embedding                      & 512-dim                                 \\
                                    & HGT hidden dimension                & 128                                     \\
                                    & HGT attention heads                 & 4                                       \\
                                    & HGT layers                          & 2                                       \\
        \hline
        \multicolumn{3}{l}{\textit{Policy core}}                                                                    \\
                                    & Concatenated state vector w/ depth  & 1888-dim                                \\
                                    & Concatenated state vector w/o depth & 1632-dim                                \\
                                    & LSTM layers                         & 2                                       \\
                                    & LSTM hidden units                   & 1024                                    \\
        \hline
        \multicolumn{3}{l}{\textit{Action space --- Multi-Head (Variant B)}}                                        \\
                                    & Rotation angles                     & 24 (0--345$^\circ$ in 15$^\circ$ steps) \\
                                    & Lengths                             & 21 (0.0--2.0\,m in 0.1\,m steps)        \\
                                    & Stop head                           & 2 (binary)                              \\
                                    & Motion resolution                   & 504                                     \\
        \hline
        \multicolumn{3}{l}{\textit{Action space --- Single-Head (Variant A)}}                                       \\
                                    & Compact action set                  & 16 (8 rots $\times$ 2 lens)             \\
                                    & Enlarged action set                 & 504 (24 rots $\times$ 21 lens)          \\
        \hline
        \multicolumn{3}{l}{\textit{Curriculum stages }}                                                             \\
                                    & Stage 1                             & 16 (8 rots $\times$ 2 lens)             \\
                                    & Stage 2                             & 48 (8 rots $\times$ 6 lens)             \\
                                    & Stage 3                             & 160 (8 rots $\times$ 20 lens)           \\
                                    & Stage 4                             & 504 (24 rots $\times$ 21 lens)          \\
        \bottomrule
    \end{tabular}
\end{table}
\begin{table}[htbp]
    \centering
    \caption{Hyperparameters for training and reward shaping.
        PPO entries listed as “SH16 / SH504 / MH” correspond to the compact single-head action space (16 actions),
        the full single-head action space (504 actions), and the multi-head factorised action space (24 rotation angles, 21 lengths, \textsc{Stop}), respectively.}
    \label{tab:hyperparameters}
    \small
    \setlength{\tabcolsep}{6pt}
    \renewcommand{\arraystretch}{1.15}
    \begin{tabular}{lll}
        \toprule
        \textbf{Category} & \textbf{Parameter}                          & \textbf{Value}                                \\
        \midrule
        \multicolumn{3}{l}{\textit{PPO hyperparameters (SH16 / SH504 / MH)}}                                            \\
                          & Learning rate $\alpha$                      & $1\text{e-}4$ / $1\text{e-}4$ / $1\text{e-}4$ \\
                          & Discount factor $\gamma$                    & 0.99                                          \\
                          & GAE lambda $\lambda$                        & 0.90 / 0.97 / 0.97                            \\
                          & Clipping parameter $\epsilon$               & 0.25 / 0.15 / 0.15                            \\
                          & Value loss coefficient $c_v$                & 0.5                                           \\
                          & Entropy coefficient $c_{\text{ent}}$        & 0.05 / 0.05 / 0.05                            \\
                          & Collision loss coefficient $c_{\text{aux}}$ & 0.2 / 0.3 / 0.3                               \\
                          & Epochs per update $K$                       & 4                                             \\
                          & Gradient clipping (max\_norm)               & 0.5                                           \\
                          & KL target (early stopping)                  & 0.01                                          \\
        \addlinespace
        \multicolumn{3}{l}{\textit{REINFORCE hyperparameters}}                                                          \\
                          & Learning rate $\alpha$                      & $5\text{e-}4$                                 \\
                          & Discount factor $\gamma$                    & 0.97                                          \\
                          & Entropy coefficient $c_{\text{ent}}$        & 0.05                                          \\
                          & Collision loss coefficient $c_{\text{aux}}$ & 0.3                                           \\
        \addlinespace
        \multicolumn{3}{l}{\textit{Reward shaping}}                                                                     \\
                          & Node weight $\lambda_{\text{node}}$         & 0.1                                           \\
                          & Visibility weight $\lambda_p$               & 0.5                                           \\
                          & Viewpoint diversity weight $\lambda_d$      & 0.001                                         \\
                          & Edge visibility $P_{\text{edge}}$           & 1.0                                           \\
                          & Time penalty $\rho$                         & 0.001                                         \\
                          & Collision penalty (event term)              & $-0.02$                                       \\
                          & Movement bonus (event term)                 & $+0.005$                                      \\
                          & Exploration bonus (event term)              & $+0.01$                                       \\
                          & Stop bonus (event term)                     & 0.05                                          \\
                          & Min.\ steps for stop bonus $K$              & 5                                             \\
                          & Visibility threshold (Node Recall)          & 0.8                                           \\

        \bottomrule
    \end{tabular}
\end{table}

\end{document}